\documentclass[journal]{AIBM}

\usepackage[numbers,sort&compress]{natbib}
\usepackage{lineno,hyperref}
\usepackage{caption}
\usepackage{graphicx, subfig}
\usepackage{stfloats}
\usepackage{booktabs}
\usepackage{threeparttable}
\usepackage{multicol}
\usepackage{multirow}
\usepackage{enumerate}
\usepackage{amsmath}
\usepackage{amssymb}
\usepackage{footnote}
\usepackage{gensymb}
\usepackage{diagbox} 

\usepackage{hyperref}
\usepackage{enumerate}
\hypersetup{
    colorlinks=true,
    linkcolor=blue,
    filecolor=gray,
    urlcolor=blue,
    citecolor=blue,
}

\modulolinenumbers[5]
\begin{document}

\captionsetup[figure]{labelfont={bf}, labelformat={default},labelsep=period,name={Fig.}}

\title{A manometric feature descriptor with linear-SVM to distinguish esophageal contraction vigor}


\DeclareRobustCommand*{\IEEEauthorrefmark}[1]{%
    \raisebox{0pt}[0pt][0pt]{\textsuperscript{\footnotesize\ensuremath{#1}}}}

\author{\IEEEauthorblockN{Jialin~Liu\IEEEauthorrefmark{a}\IEEEauthorrefmark{1},
Lu~Yan\IEEEauthorrefmark{c}\IEEEauthorrefmark{1},Xiaowei~Liu\IEEEauthorrefmark{c},Yuzhuo~Dai\IEEEauthorrefmark{a},Fanggen~Lu\IEEEauthorrefmark{d},Yuanting~Ma\IEEEauthorrefmark{e},Muzhou~Hou\IEEEauthorrefmark{a}\IEEEauthorrefmark{*}
and Zheng~Wang \IEEEauthorrefmark{a}\IEEEauthorrefmark{b}\IEEEauthorrefmark{*}
        }

\IEEEauthorblockA{\IEEEauthorrefmark{a} School of Mathematics and Statistics, Central South University, Changsha 410083, China; }

\IEEEauthorblockA{\IEEEauthorrefmark{b} Science and Engineering School, Hunan First Normal University, Changsha 410205, China; }

\IEEEauthorblockA{\IEEEauthorrefmark{c} Department of Gastroenterology of Xiangya Hospital, Central South University, Changsha 410008, China; }

\IEEEauthorblockA{\IEEEauthorrefmark{d} The second Xiangya Hospital, Central South University, Changsha 410083, China. }

\IEEEauthorblockA{\IEEEauthorrefmark{e} Fang Zongxi Center for Marine Evo-Devo and MOE Key Laboratory of Marine Genetics and Breeding, College of Marine Life Sciences, Ocean University of China, Qingdao 266003, China. }

\thanks{
\IEEEauthorrefmark{*}Corresponding author

Email addresses: houmuzhou@sina.com (Muzhou~Hou), zhengwang@csu.edu.cn (Zheng~Wang) }
\thanks{\IEEEauthorrefmark{1}Jialin~Liu and Lu~Yan contributed equally to this work.}
}

\markboth{ }%
{Shell \MakeLowercase{\textit{et al.}}: Bare Demo of IEEEtran.cls for IEEE Journals}

\maketitle

\begin{abstract}
In clinical, if a patient presents with nonmechanical obstructive dysphagia, esophageal chest pain, and gastro esophageal reflux symptoms, the physician will usually assess the esophageal dynamic function. High-resolution manometry (HRM) is a clinically commonly used technique for detection of esophageal dynamic function comprehensively and objectively. However, after the results of HRM are obtained, doctors still need to evaluate by a variety of parameters. This work is burdensome, and the process is complex. We conducted image processing of HRM to predict the esophageal contraction vigor for assisting the evaluation of esophageal dynamic function. Firstly, we used Feature-Extraction and Histogram of Gradients (FE-HOG) to analyses feature of proposal of swallow (PoS) to further extract higher-order features. Then we determine the classification of esophageal contraction vigor normal, weak and failed by using linear-SVM according to these features. Our data set includes 3000 training sets, 500 validation sets and 411 test sets. After verification our accuracy reaches 86.83\%, which is higher than other common machine learning methods.
\end{abstract}

\section{Introduction}
\begin{figure}[h]
  \centering
  \includegraphics[width=0.5\textwidth]{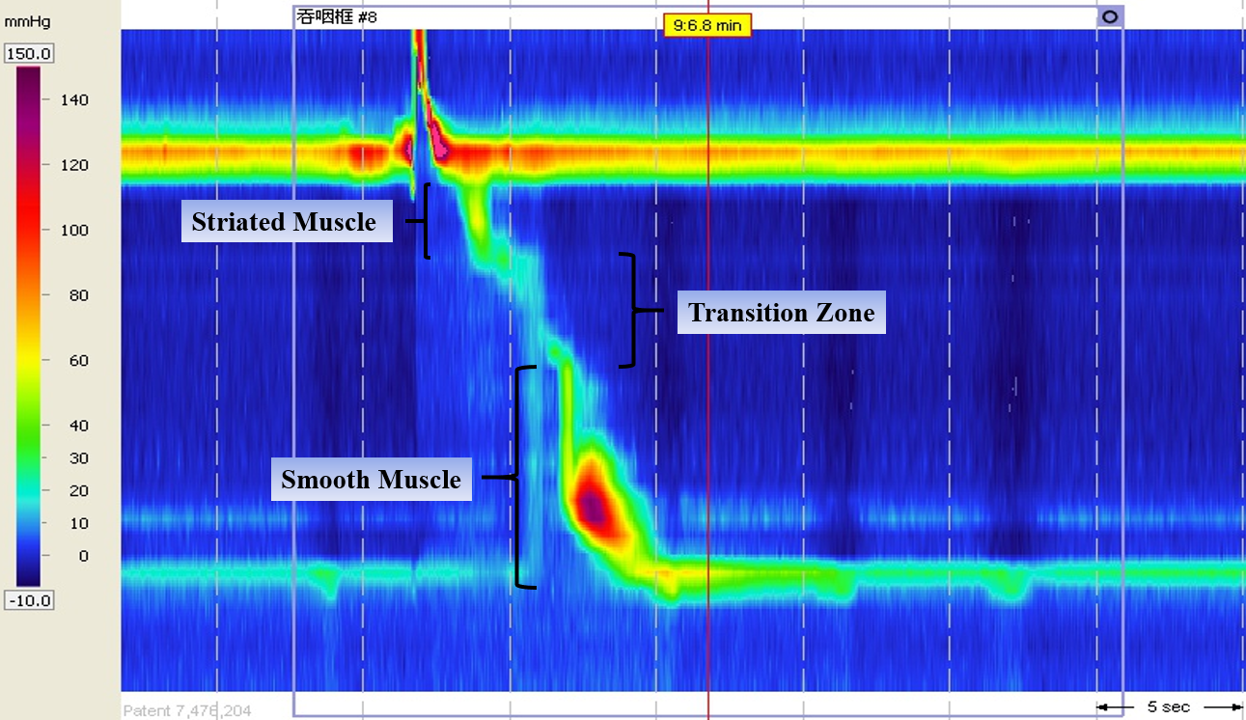}
  \caption{{\bf The HRM swallowing analysis interface image} \newline The topographic plot of high-resolution manometry is obtained by the HRM catheter and recording system, in which sensor location is on the y-axis, time is on the x-axis, and pressure is represented by color (Cooler colors represent lower pressures and warmer colors represent higher pressures)  }
  \label{fig1}
\end{figure}

\IEEEPARstart{P}{urpose} of review High-resolution manometry (HRM) is increasingly performed worldwide, to study esophageal motility \cite{Carlson2018High, Vettori2018Esophageal}. When a patient is referred with dysphagia, an upper endoscopy is the essential first step to exclude structural abnormalities such as esophageal carcinoma, stricture, or eosinophilic esophagitis. If no abnormalities are found, we also consider upper gastrointestinal barium meal. When they both have no problem, esophageal manometry is usually the next step for the detection of esophageal motility disorders. Therefore, HRM is now considered the gold standard for diagnosis of esophageal motility disorders\cite{RohofChicago}. HRM is obtained by pressure sensors spaced at 1cm intervals from the pharynx to the stomach. HRM reveals the segmental character of oesophageal peristalsis and the functional anatomy of the esophago-gastric junctiong (EGJ). Even more fundamentally, HRM displays the pressure gradient during bolus transport, the force that directs the movement of food and fluid through the oesophagus and into the stomach \cite{BredenoordHigh}. The HRM swallowing analysis interface image is shown as Figure \ref{fig1}.

HRM is usually performed in conjunction with the Chicago classification system. The initial step is to evaluate the relaxation of the esophagogastric junction upon swallowing by using the integrated relaxation pressure (IRP). If elevated, patients should be classified as having achalasia or EGJ outflow obstruction, depending on the peristalsis. In case of a normal IRP, peristalsis is classified based on distal latency, distal contractile integral (DCI), and fragmentation. If there are abnormalities, patients are classified as having a major or minor disorder of peristalsis. Major disorders are never observed in controls, in contrast to minor disorders. If a patient has a normal IRP and more than 50\% of swallows are effective, esophageal motility is normal\cite{RohofChicago}. It can be seen that the process of diagnosing HRM through the Chicago classification is burdensome and the parameters are complex. During the evaluation, we found that the esophageal contraction vigor is an important indicator to assist the doctor's diagnosis.

Machine learning can be traced back to the research of artificial neural network \cite{Mcculloch1943A}. At present, machine learning algorithms include classification algorithm \cite{Kim2012A, FrancoBuilding}, clustering algorithm \cite{HavensFuzzy, Havens2011Speedup} and feature selection algorithm \cite{5419028, SunLocal}. In recent years, machine learning has made amazing achievements in many fields, such as image processing \cite{Ibrahim2018Fish, Xuegang2019Object, wang}, nature language processing \cite{article, article2}, Internet security \cite{Sherali2019Securing, article3} and fitting prediction \cite{article4, Heng2019High}.
So far, there are many studies focusing on medical images process using machine learning. Line operators and support vector classification are used to segment retinal blood vessel \cite{4336179}. J. Zhang, Y.Z.Gao, and et.al. proposed segmentation of perivascular spaces using vascular features and structured random forest from 7T MR image \cite{Zhang2016Segmentation}. In 2019, Neffati classified brain MR images by enhanced SVM-KPCA method \cite{10.1093/comjnl/bxz035}. Machine learning methods also be used for radiomics-based differentiation of local recurrence versus inflammation from PET/CT images\cite{Du2019Machine}. 

In this paper, we propose to recognize the esophageal contraction vigor (normal, weak and failed) directly and aotomatic by the machine learning methods of FE-HOG and Linear-SVM through the processing of HRM images. There are three main highlights of this paper:
\begin{enumerate}
\item  In this paper, the machine learning method was proposed to assist the evaluation of esophageal dynamic function for the first time, which offered new thoughts for the application of machine learning in the field of esophageal pressure measurement. This was conducive to further development.
\item In the method, feature extraction is added to the feature analysis, making the accuracy significantly higher than the traditional HOG method.
\item We compared Liner-SVM with several commonly used machine learning methods (Nearest Neighbors, RBF SVM, Decision Tree, Random Forest and AdaBoost), and the results were significantly better than other methods. Our accuracy reaches 86.83\%.
\end{enumerate}

\section{material and methods}
\begin{figure}[hb]
\setlength{\belowcaptionskip}{-0.5cm}
  \centering
  \includegraphics[width=0.5\textwidth]{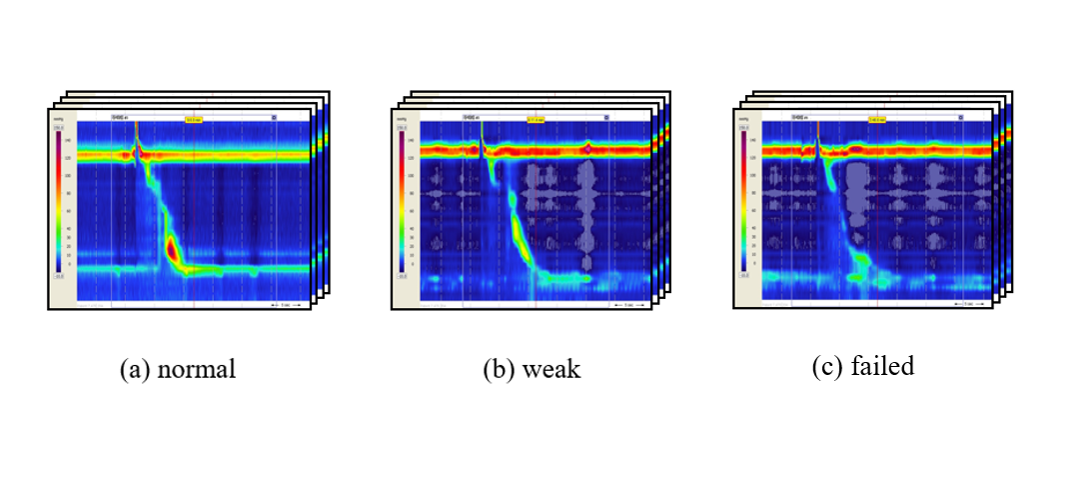}
  \caption{{\bf the dataset used for the experiments} }
  \label{fig2}
\end{figure}

Our dataset is based on the image data collection and restrospective study protocol approved by XiangYa Hospital of Central South University. The data used for the experiments were detected by ManoScan 360TM (American Sierra Scientific Instruments), a solid-state high-resolution esophageal pressure measurement system, and analyzed by Mano View ESO 3.0 software. Our dataset are shown as Figure \ref{fig2}, including 1500 normal HRM, 1050 weak HRM and 1450 failed HRM. Since the data we collected was missing from the hyperconractile category, we categorized only abnormal, weak, and failed in this experiment.

\subsection{Overall Framework}
\begin{figure*}[ht]
\setlength{\belowcaptionskip}{-0.5cm}
  \centering
  \includegraphics[width=1\textwidth]{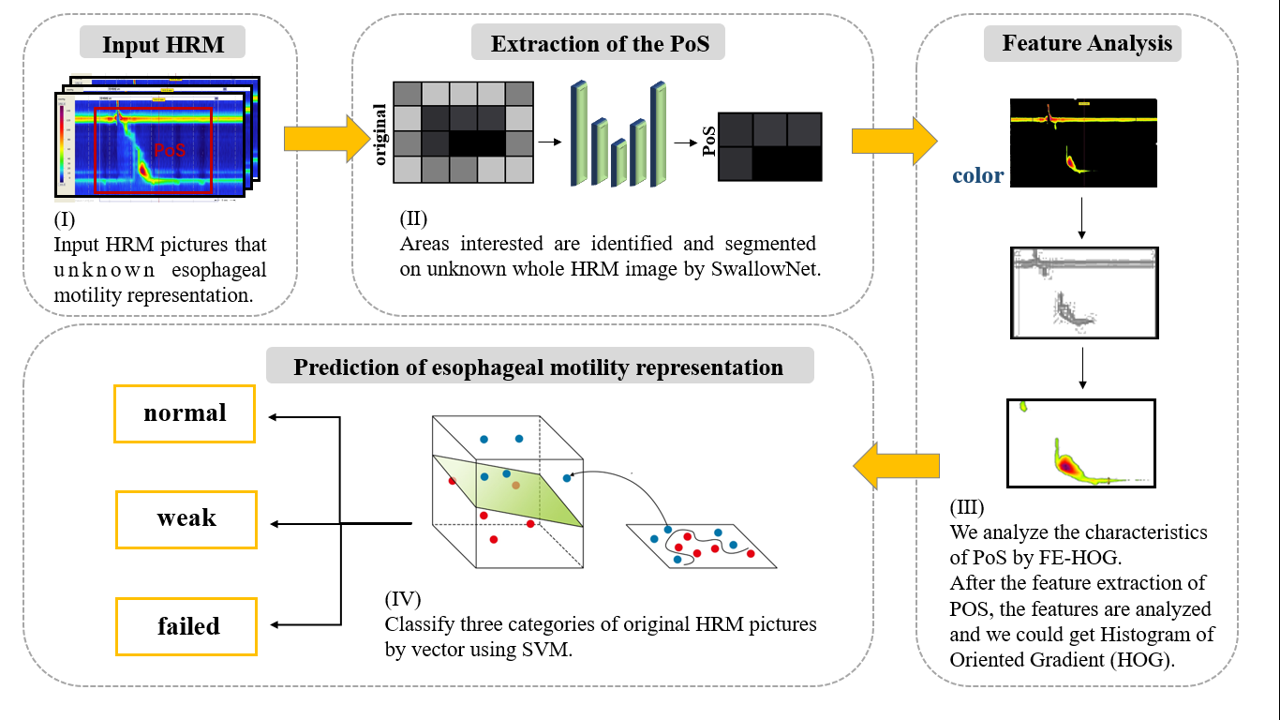}
  \caption{{\bf the picture of model frame} \newline For any input HRM image, we first crop the PoS from the image to remove the background by using U-Net. Then we extract feature extraction and feature processing on the cropped image. In this setp, we could get HOG, which represents the gradient of each pixel in the image. After the feature vector of the image is calculated, it is input into the SVM for training and recognition. Finally, the SVM will automatically recognize the esophageal contraction vigor of the input HRM picture.}
  \label{fig3}
\end{figure*}

The flow chart of our method is illustrated in Figure \ref{fig3}. In the first phase, we selelcte proposal of swallow (PoS) under the gastroenterologist help on a whole HRM image. These are cropped into smaller slices, which are fed into the SwallowNet matching the original images as training examples. Then HRM picture is input into the trained SwallowNet network, the PoS of this picture will be output automatically. Next PoS image is sent to the second step, which is prcessing extract analysis by FE-HOG aogorithm. In this step, our FE-HOG algorithm is improved on the based of traditional HOG algorithm for extracting features better, which could contribute to subsequent prediction. The feature vectors extracted by HOG algorithm were input into SVM for training and recognition, and the optimal parameters were found.Finally, the trained SVM classifier can predict the esophageal contraction vigor of the original HRM images.

 In the following sections, we will provide a detailed description of how to extract the PoS, analyse the feature and predict the esophageal motility representation.

\subsection{Extraction of the PoS  -- Swallowing box segmentation}
As we can see in the Figure \ref{fig2}, HRM images cover a wide range of areas, including swallowing box and non-swallowing parts. Obviously, the non-swallowing parts are meaningless to our prediction, and when we focus on the swallowing box, we can make distinguish more quickly and accurately. So the first step, swallowing box segmentation, is one of key techniques in computer aided diagnosis systems to distinguish esophageal contraction vigor. Inspired from U-Net we developed featureaware manometric descriptor, SwallowNet, similar to encoder-decoder structure.
\begin{enumerate}
\item contracting path
\newline After the input of the initial image, we under-sampled the resolution and transmitted the feature map through four layers of convolution. Here, repeated convolution is used to generate multi-scale feature maps that reconstruct the resolution abundant in semantic information in a top-to-down path. We all used 3 $*$ 3 convolution kernel, thepadding is 0, and the striding is 1. After each convolution is a Max pooling with a stride of 2. The last time there is no Max pooling, we directly send the obtained feature map to the Decoder.
\item expanding path
\newline Next, the extended decoder is used to up-sampled, and the local context information is propagated in parallel with the contraction encoder, that is, the location information of the underlying information and the semantic information of the deep features are fused, and the original resolution is finally restored. This pattern preserves more dimensional information and makes semantic segmentation more accurate.
\end{enumerate}

\subsection{Feature Analysis - feature-aware manometric descriptor}

\begin{figure}[h]
\setlength{\belowcaptionskip}{-0.5cm}
  \centering
  \includegraphics[width=0.5\textwidth]{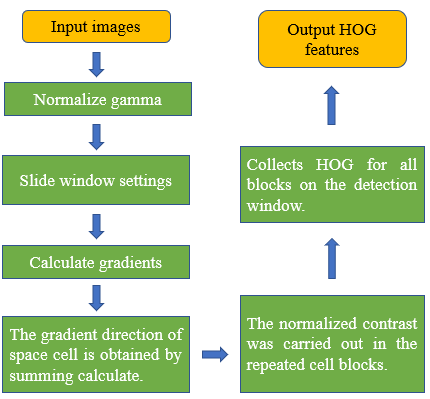}
  \caption{{\bf feature-aware manometric descriptor} }
  \label{fig4}
\end{figure}

High-resolution manometry (HOG) is a kind of feature descriptor, proposed by Navneet and Bill in 2005 \cite{HOG}. Feature descriptors can better recognize the edge information of an object and thus get the shape, which is helpful for figure detection. In this section, we will go into the details of calculating the HOG feature descriptor. The flow chart of HOG feature extraction is described by Figure \ref{fig4}.

\begin{itemize}
\item[-] We use gamma transform to normalize the image for resize the contrast of image that can effectively reduce the local shadow and light changes in the image, in the same time weaken the noise interference.
\item[-] We divide the image into blocks: each Block slides over the image at a certain step size to create a new Block. As the basic feature extraction unit, Block is subdivided again. Divide the Block into (usually evenly divided) N*N pieces, each of which is called a cell.
\item[-] The cell is the most basic statistical unit.Inside the cell, we calculate the horizontal and vertical gradients for each pixel (x, y).
\newline
The horizontal gradient is written as
\begin{equation}
|_{x}(x, y)=|(x+1, y)-|(x-1, y)
\end{equation}
\newline
The vertical gradient is written as
\begin{equation}
|_{y}(x, y)=|(x, y+1)-|(x, y-1)
\end{equation}
\newline
So, the gradient of pixel (x, y) is
\begin{equation}
m(x, y)=\sqrt{\left(I_{x}(x, y)\right)^{2}+\left(I_{y}(x, y)\right)^{2}}
\end{equation}
\newline
In the same way, we could calculate the gradient direction of pixel (x, y)
\begin{equation}
\theta(x, y)=\arctan \left(\frac{I_{y}(x, y)}{I_{x}(x, y)}\right)
\end{equation}
\item[-] Then the gradient direction of the space cell is calculated by summing.Divide the gradient direction evenly into m directions (bin). If the gradient direction is positive or negative, divide 360\degree evenly into m intervals; otherwise, divide 180\degree evenly into m intervals.The point gradient amplitudes of all the same gradient directions on the same cell were calculated based on weight accumulation to obtain the HOG of the cell.
\item[-] Then we combine gradient histograms of multiple cell within each block into one histogram to represent the HOG feature of the current block.
\item[-] The HOG feature of the whole image is extracted by sliding block window.
\item[-] In the end, the HOG feature is output, that is, the final feature vector of the whole image.
\end{itemize}

Since the classifier usually takes up a large space and the training speed is slow, in order to reduce the pressure of classification, we consider to remove the useless pixels, that is, the pixels that have no effect on the recognition of swallowing function. In our dataset of HRMs, colors are assigned to pressures. Cooler colors represent lower pressures and warmer colors represent higher pressures. We improved the HOG due to the color feature of HRM image, which named Highresolution manometry with Feature Extraction (FE-HOG). 

We set up a color list based on the color feature of dataset as Table \ref{table1}. The color list is used to match the color to the pressure value. We let the RGB of lower green is [35, 43, 46], the upper green is [45, 255, 255], this is the threshold for pressure in potential color range.

\begin{table}[h]
  \center
  \fontsize{10}{18}\selectfont
  \captionsetup{labelfont=bf, singlelinecheck=off}
  \caption{\newline The color list}
  \label{table1}
  \setlength{\tabcolsep}{12mm}{
  \begin{threeparttable}
    \begin{tabular}{c|c}
    \cmidrule(lr){1-2}
    color & The RGB value \cr
    \midrule
    lower red	& [156, 43, 46] \cr
    upper red	& [180, 255, 255] \cr
    lower red	& [0, 43, 46] \cr
    upper red	& [10, 255, 255] \cr
    lower orange &	[11, 43, 46] \cr
    upper orange &	[25, 255, 255] \cr
    lower yellow &	[26, 43, 46] \cr
    upper yellow &	[34, 255, 255] \cr
    \bottomrule
    \end{tabular}
  \end{threeparttable}}
\end{table}

The results of feature extraction using HOG and fe-hog are input into the classifier, and the obtained precision results are shown in Table \ref{table2} by the example of Linear SVM method.

\begin{table}[h]
  \center
  \fontsize{10}{18}\selectfont
  \captionsetup{labelfont=bf, singlelinecheck=off}
  \caption{\newline The comparison of HOG and FE-HOG}
  \label{table2}
  \setlength{\tabcolsep}{4mm}{
  \begin{threeparttable}
    \begin{tabular}{c|cc}
    \cmidrule(lr){1-3}
    methods & HOG & FE-HOG \cr
    \midrule
    the accuracy of classification & 75.61\% & 86.83\% \cr
    \bottomrule
    \end{tabular}
  \end{threeparttable}}
\end{table}

In Table 1, we could find that after feature extraction, HOG greatly improves the classification accuracy.

\subsection{Classification by SVM - machine learning}
SVM is a machine learning algorithm based on the minimization of structural risk, which is essentially a kernel method. It is usually used to solve binary classification problems, but can also be used to solve multi-classification problems. It is very effective for solving nonlinear, small sample, high dimension and local minimum problems \cite{Support}. SVM is a supervised learning algorithm, which is usually used to analyze the linear separable problem. The problem is transformed into the linear separable problem in the high dimensional space to construct the optimal classification surface. For the binary classification problem, since the segmentation hyperplane can "tolerate" the local interference of the training samples, it is necessary to find the segmentation hyperplane with the best "tolerance" performance which is located in the middle of the two types of training samples.

The learning process of SVM is as follows:
\begin{enumerate}
\item Feature extraction was performed on the test set of the sample set.
\item Select the appropriate kernel function for transformation, and convert the input sample into a high-dimensional space.
\item Construct the optimal separation hyperplane in the high dimensional space, that is, search SVM. The SVM was used to construct the learning machine and complete the training of the samples.
\item Input the unknown data of the same preprocessing into the learning machine for classification and discrimination, and get the learning result. The learning process is over.
\end{enumerate}

In this experiment, the SVM model of sklearn in python was used for processing, and the penalty parameter c with the highest accuracy was obtained after cross validation. Then, the SVM model parameters were obtained by training the whole training set.The kernel function chosen by experiment is linear kernel function  \begin{equation}K\left(\mathbf{x}_{i}, \mathbf{x}_{j}\right)=\mathbf{x}_{i}^{T} \mathbf{x}_{j}\end{equation} and the penalty parameter for the error term is 0.025.

\section{Overview of Results}
In this section, we built the framework to run a series of experiments by using python. For our empirical evaluation of our learning methods we used a 4000 three-category image dataset from XiangYa Hospital of Central South University, including normal, weak and failed.
We divided our experiments into two main parts:

In part one, we first obtained the swallowing box by SwallowNet. Then we removed the useless pixels in the swallowing box with the color list by using FE-HOG, the useless pixels are the pixels independent of esophageal contraction vigor. We also input the results processed by Fe-HOG and HOG into different classifiers for comparison, and the advantages of FE-HOG can be intuitively found.

In part two, we trained different machine learning classifiers to classify our data. By comparing the classification results of different classifiers, we can prove that SVM is significantly better than other machine learning methods.

\subsection{Evaluation indicators}
First of all, Let's be clear about the Common model evaluation terms before the introduction of evaluation indicators. In our experiments, we will devide the dataset into three categories. For each category, we treat it as positive and the other two as negative. 
\begin{enumerate}
\item True positives (TP) represents the number of samples that are correctly classified as positive, that is, the number of samples that are actually positive and are classified as positive by the classifier; 
\item False positives (FP) represents the number of samples incorrectly classified as positive, that is, the number of samples actually negative but classified as positive by the classifier
\item False negatives (FN) represents the number of samples incorrectly classified as negative, that is, the number of samples actually positive but classified as negative by the classifier
\item True negatives (TN) represents the number of samples that are correctly classified as negative, that is, the number of samples that are actually negative and are classified as negative by the classifier
\end{enumerate}

{\bf Accuracy} 
\newline Accuracy is the most common evaluation indicator, it's easy to understand that indict the number of samples classified exactly devided by the amount of samples.
\begin{equation}
accuracy = \frac{TP+TN}{P+N}
\end{equation}

{\bf Precision} 
\newline Precision is a measure of the degree of accuracy, indicating the proportion of a sample divided into positive,
\begin{equation}
precision = \frac{TP}{TP+FP}
\end{equation}

{\bf Recall} 
\newline Recall is a measure of coverage to calculate the proportion of correct classification in the positive example.
\begin{equation}
recall = \frac{TP}{TP+FN} 
\end{equation}

{\bf f1-score}
\newline $\mathrm{f}_{\mathrm{b}}$ is a harmonic average of accuracy and recall rates

\begin{equation}
\frac{\left(1+\mathrm{b}^{2}\right) * \mathrm{Precision} * \mathrm{Recall}}{\left(\mathrm{b}^{2} * \mathrm{Precision}+\mathrm{Recall}\right)}
\end{equation}

In this experiment, we choose the most common indicator f1-score.

\subsection{Feature analysis}
\begin{figure}[hb]
\setlength{\belowcaptionskip}{-0.5cm}
  \centering
  \includegraphics[width=0.5\textwidth]{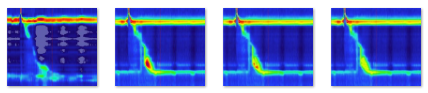}
  \caption{{\bf the swallowing box} }
  \label{fig5}
\end{figure}

After swallowing box segmentation by SwallowNet, we get the dataset of swallowing box like Figure \ref{fig5}. Based on the relationship between image colors and esophageal contraction vigor, we made feature analysis by the FE-HOG. We convert the swallowing box to HOG feature data, and then input them into the classifier. In this part, we used 4 classifier to compare the HOG and FE-HOG. There are common machine learning classifier including Decision Tree, Random Forest, AdaBoost and Linear SVM. The decision tree adopts 5 layers as the optimal value of the maximum depth tree. The maximum depth tree and the number of trees in the random forest were set as 5 and 10, respectively. AdaBoost choosed decision tree as base classifier.

The results are shown in Table \ref{table3}, and the bold values indicate results that are more efficient or accurate under the same classifier condition. In Decision Tree, FE-HOG is nearly 100 seconds faster and 20\% more accurate than HOG. In Random Forest, FE-HOG is slightly faster than HOG and 20\% more accurate. In AdaBoost, FE-HOG is about 1,000 seconds faster than HOG and 16\% more accurate. In Linear-SVM, FE-HOG is nearly 600 seconds faster and 11\% more accurate than HOG. It's clear that FE-HOG is better than HOG in efficiency and accuracy.

\begin{table*}
\center
  \fontsize{10}{18}\selectfont
  \captionsetup{labelfont=bf, singlelinecheck=off}
  \caption{\newline The comparison of HOG and FE-HOG}
  \label{table3}
  \setlength{\tabcolsep}{6mm}{
  \begin{threeparttable}
    \begin{tabular}{c|c|c|c|c}
    \cmidrule(lr){1-5}
    \multirow{2}*{methods} & \multicolumn{2}{c|}{FE-HOG} & \multicolumn{2}{c}{HOG} \cr
    ~ & time(s) & the accuracy & time(s) &	the accuracy \cr
    \midrule
    Decision Tree & {\bf 34.3754} & {\bf 0.7098} & 133.9905 & 0.5000 \cr
    Random Forest & {\bf 2.5588} & {\bf 0.6341} & 2.6581 & 0.4463 \cr
    AdaBoost & {\bf 372.9583} & {\bf 0.8073} & 1428.8249 & 0.6488 \cr
    Linear SVM & {\bf 727.6761} & {\bf 0.8683} & 1304.1814 & 0.7561 \cr

    \bottomrule
    \end{tabular}
  \end{threeparttable}}
\end{table*}

\subsection{Comparison of different classifiers}
To measure classification performance, we input the results of feature extraction by FE-HOG into six types of machine learning classifiers. At each run, we set the data in a ratio of 7:2:1 for training set, verification set and test set.

We trained Linear-SVM with 2800 groups of characteristic data, and established a model for HRM image automatic recognition.At the same time, we also realized some classifiers to compare with Linear-SVM through scikit-Learn library in Python, including Nearest Neighbors, RBF SVM, Decision Tree, Random Forest and AdaBoost. Same as PART B, the optimal value of the maximum depth tree in decision tree is 5 layers. The maximum depth tree and the number of trees in the random forest were 5 and 10. AdaBoost uses the decision tree as the base classifier, and the number of base learners is set to 10. In addition, the other parameters take default values.

The results by 6 classifiers presented in Table \ref{table4}, and the bold numbers represent the best performance. Obviously, the Linear-SVM classifier used in this paper has the best performance, the accuracy reached 86.83\%, the precision is 0.87, the recall is 0.87 and f1-score is 0.87. It got top marks at accuracy, precision, recall and f1-score. In addition, the detailed results are shown in Table \ref{table5}.

\begin{table*}[h]
  \center
  \fontsize{10}{18}\selectfont
  \captionsetup{labelfont=bf, singlelinecheck=off}
  \caption{\newline The comparison of different classifiers}
  \label{table4}
  \setlength{\tabcolsep}{6mm}{
  \begin{threeparttable}
    \begin{tabular}{c|cccc}
    \cmidrule(lr){1-5}
    classifiers & accuracy & precision & recall & f1-score \cr
    \midrule
    Nearest Neighbors	 & 0.4195 & 0.61 & 0.42 & 0.32 \cr
    RBF SVM & 0.7122 & 0.72 & 0.71 & 0.67 \cr
    Decision Tree & 0.7098 & 0.72 & 0.71 & 0.7 \cr
    Random Forest & 0.6341 & 0.5 & 0.63 & 0.54 \cr
    AdaBoost & 0.8073	 & 0.82 & 0.81 & 0.81 \cr
    Linear SVM & {\bf 0.8683} & {\bf 0.87} & {\bf 0.87}	 & {\bf 0.87} \cr

    \bottomrule
    \end{tabular}
  \end{threeparttable}}
\end{table*}

\begin{table*}[h]
  \center
  \fontsize{10}{18}\selectfont
  \captionsetup{labelfont=bf, singlelinecheck=off}
  \caption{\newline The results by Linear-SVM}
  \label{table5}
  \setlength{\tabcolsep}{6mm}{
  \begin{threeparttable}
    \begin{tabular}{c|cccc}
    \cmidrule(lr){1-5}
      & precision & recall & f1-score & support \cr
    \midrule
    normal & 0.86 & 0.95 & 0.90 & 148 \cr
    weak & 0.92 & 0.91 & 0.92 & 155 \cr
    failed & 0.80 & 0.69 & 0.74 & 107 \cr
    total	 & 0.87 & 0.87 & 0.87	 & 410 \cr
    \bottomrule
    \end{tabular}
  \end{threeparttable}}
\end{table*}

\section{Conclusions}
In this article, we applied machine learning method to distinguish esophageal contraction vigor by HRM for the first time. This method could greatly reduce the workload of doctor's diagnosis, and promoted further development.

In addition, we improved the traditional HOG method to analysis feature. We use the HRM images color feature, the corresponding relationship between color and esophageal contraction vigor, to set up color list. Then we removed useless pixels by FE-HOG to greatly improving the identification accuracy and speeding up the operation speed.

Finally, it is proved that Linear-SVM is the best classifiervby comparing Linear-SVM with common machine learning classifier.

In the future, we will try to further improve the performance of esophageal contraction vigor prediction, and we also try to use machine learning methods for in-depth diagnosis to assist doctor's work.

\section{Compliance with Ethical Standards}
\begin{itemize}
\item[-] {\bf Conflict of interest} The authors declare that they have no conflict of interest.

\item[-] {\bf Ethical approval} All procedures performed in studies involving human participants were in accordance with the ethical standards of the Institutional Ethics Committee of XiangYa Hospital of Central South University and with the 1964 Helsinki Declaration and its later amendments or comparable ethical standards. 

\item[-] {\bf Informed consent} Additional informed consent was obtained from all individual participants.
\end{itemize}

\newpage
\bibliography{FE-HOG}                    
\bibliographystyle{IEEEtran}

\end{document}